\documentclass[10pt,twocolumn,letterpaper]{article}

\usepackage[pagenumbers]{iccv} 

\usepackage{microtype}
\usepackage{graphicx}

\usepackage{amsmath}
\usepackage{amssymb}
\usepackage{mathtools}
\usepackage{amsthm}
\usepackage{multirow}

\usepackage{pifont}
\usepackage{float}  
\usepackage{algorithm}
\usepackage{algorithmic}

\newcommand{\cmark}{\ding{51}} 
\newcommand{\xmark}{\ding{55}} 

\definecolor{iccvblue}{rgb}{0.21,0.49,0.74}
\usepackage[pagebackref,breaklinks,colorlinks,allcolors=iccvblue]{hyperref}

\def\confName{arXiv}
\def\confYear{2026}

\title{MEDiC: Multi-objective Exploration of Distillation from CLIP}

\author{
    Konstantinos Georgiou
    \and
    Maofeng Tang
    \and
    Hairong Qi\\
    \and
    Min H. Kao Department of Electrical Engineering and Computer Science\\
    The University of Tennessee\\
    Knoxville, TN 37996\\
    {\tt\small \{kgeorgio, mtang6\}@vols.utk.edu, hqi@utk.edu}
}

\begin{document}
\maketitle
\begin{abstract}
Masked image modeling (MIM) methods typically operate in either raw pixel space (reconstructing masked patches) or latent feature space (aligning with a pre-trained teacher). We present MEDiC (Multi-objective Exploration of Distillation from CLIP), a framework that combines both spaces in a single pipeline through three complementary objectives: patch-level token distillation from a frozen CLIP encoder, global CLS alignment, and pixel reconstruction via a lightweight decoder. We conduct a systematic investigation of the design space surrounding this multi-objective framework. First, we show that all three objectives provide complementary information, with the full combination reaching 73.9\% kNN accuracy on ImageNet-1K. Second, we introduce hierarchical clustering with relative position bias for evolved masking and find that, despite producing more semantically coherent masks than prior methods, evolved masking does not outperform simple block masking in the teacher-guided distillation setting, a finding we attribute to the teacher's inherent semantic awareness. Third, we reveal that optimal scalar loss weights are extremely fragile, with small perturbations causing drops of up to 17 percentage points in kNN accuracy. Our framework achieves 73.9\% kNN and 85.1\% fine-tuning accuracy with ViT-Base at 300 epochs.
\end{abstract}

\begin{figure*}[t!]
    \centering
    \includegraphics[width=\textwidth]{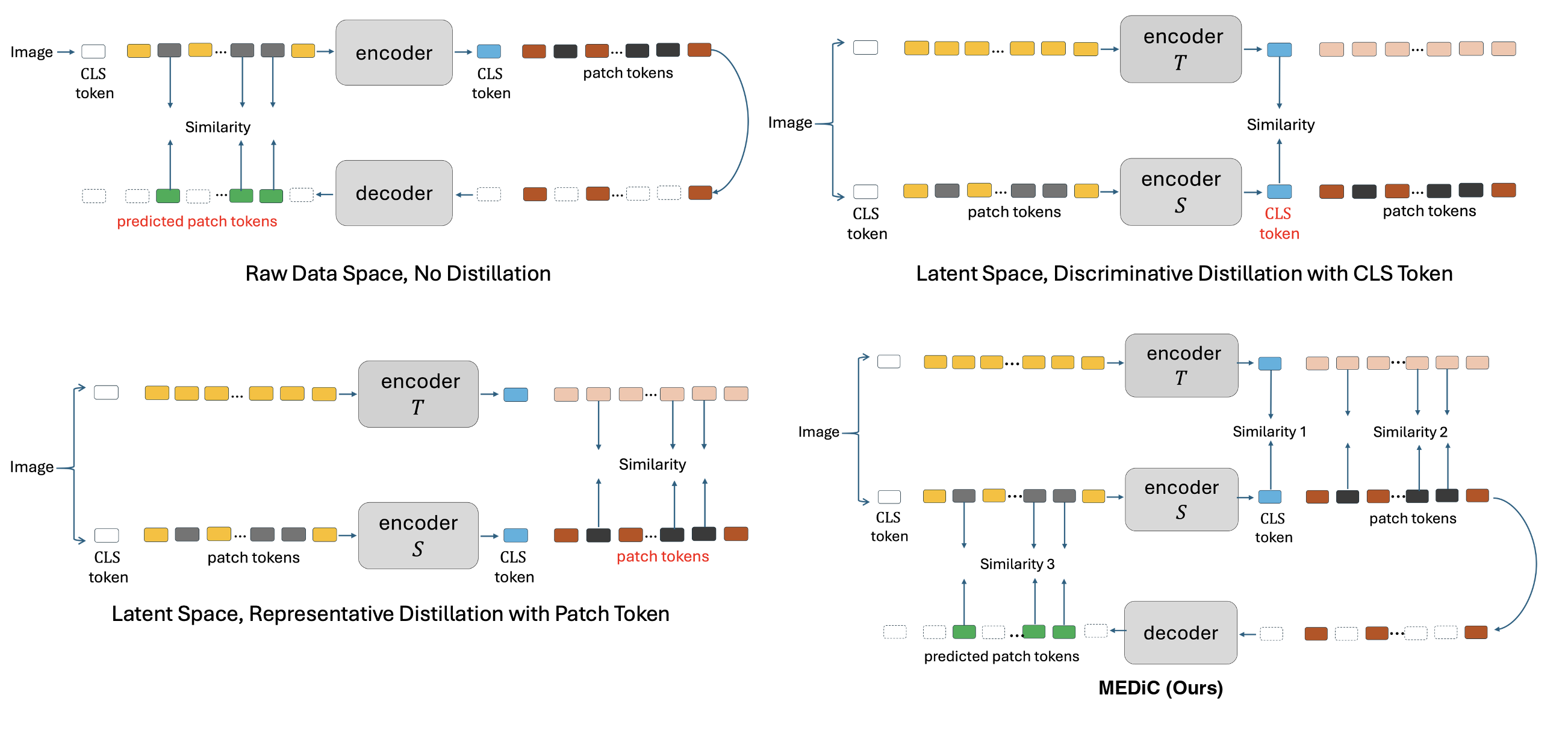}
    \caption{Four paradigms in masked image modeling. Top-left: raw-space pixel reconstruction (MAE-style). Top-right: latent-space prediction with discrete visual tokens (BEiT-style). Bottom-left: latent-space distillation at the patch level from a teacher (MaskDistill-style). Bottom-right: MEDiC combines pixel reconstruction with both patch-level and CLS-level distillation from a frozen CLIP teacher.}
    \label{fig:MEDiC}
\end{figure*}

\section{Introduction}

Masked image modeling (MIM) has established itself as a leading paradigm for self-supervised visual representation learning~\cite{he2022masked, bao2021beit, xie2022simmim}. At its core, MIM masks a portion of image patches and trains the model to predict the missing content, forcing the encoder to develop representations that capture both local structure and global context. A central design question in MIM is what the model should predict for masked patches: raw pixel values~\cite{he2022masked, xie2022simmim}, discrete visual tokens from a learned codebook~\cite{bao2021beit, peng2022beit}, or latent features from a pre-trained teacher such as CLIP~\cite{radford2021learning, dong2022bootstrapped, hou2022milan}.

Each reconstruction target captures different aspects of visual information. Pixel-level reconstruction preserves fine-grained spatial detail but may overemphasize low-level texture. Teacher-based distillation at the patch level transfers rich semantic features but may neglect local nuances that the teacher's representations discard. Global alignment through classification (CLS) tokens ensures image-level coherence but provides no patch-level learning signal. A natural question arises: can these complementary objectives be combined in a single framework to capture information at multiple levels simultaneously?

In this work, we present MEDiC (Multi-objective Exploration of Distillation from CLIP), a framework that addresses this question through systematic investigation of multi-objective masked distillation. MEDiC operates in both raw data space and latent feature space, combining three complementary objectives: (1)~patch-level token distillation that aligns student representations with a frozen CLIP teacher, (2)~global CLS alignment that preserves image-level semantics, and (3)~pixel reconstruction through a lightweight decoder that grounds the representation in raw visual content.

Beyond the multi-objective framework itself, we investigate two additional design dimensions that interact with multi-objective distillation. First, we explore whether sophisticated masking strategies can further improve representations. We introduce hierarchical clustering (HC) with relative position bias for evolved masking, which produces more semantically coherent mask patterns than prior expectation-maximization approaches~\cite{feng2023evolved}. However, we find that in the CLIP-distillation setting, even the best evolved masking configuration does not surpass simple block masking---a result we attribute to the teacher already providing semantic guidance that overlaps with what attention-guided masking attempts to achieve. Second, we conduct a comprehensive analysis of loss weight sensitivity, revealing that scalar weights are extremely fragile: the optimal pixel reconstruction weight (0.01) yields 71.4\% kNN accuracy, while a seemingly reasonable value of 0.50 drops performance to 61.6\%, a difference of nearly 10 points across a narrow range.

Our contributions are:
\begin{itemize}
    \item[(1)] A multi-objective framework that combines pixel reconstruction, patch-level CLIP distillation, and global CLS alignment, demonstrating that all three objectives provide complementary information with the full combination outperforming any subset.
    \item[(2)] An improved evolved masking strategy using hierarchical clustering with relative position bias, which produces more coherent masks than EM-based approaches, along with the finding that block masking remains superior in teacher-guided distillation settings.
    \item[(3)] A systematic analysis of loss weight sensitivity and dense versus sparse encoding, revealing that scalar loss weights are extremely fragile and that sparse encoding consistently outperforms dense encoding for multi-objective MIM.
    \item[(4)] Strong results on ImageNet-1K with ViT-Base at 300 epochs: 73.9\% kNN, 85.1\% fine-tuning accuracy, and competitive downstream performance.
\end{itemize}

\section{Related Work}
\label{sec:related_work}

\textbf{Masked Image Modeling.}
Masked image modeling (MIM) adapts the masked prediction paradigm from NLP~\cite{devlin2019bert} to vision. BEiT~\cite{bao2021beit} introduced block-wise masking with a pre-trained dVAE tokenizer to predict discrete visual tokens. MAE~\cite{he2022masked} proposed an asymmetric encoder-decoder that masks 75\% of patches and reconstructs raw pixels, demonstrating that high masking ratios with simple pixel targets can learn strong representations. SimMIM~\cite{xie2022simmim} showed that direct pixel prediction with large square patches is competitive with more complex targets. iBOT~\cite{zhou2021ibot} combined masked prediction with self-distillation using a momentum teacher, bridging MIM with the DINO~\cite{caron2021emerging} paradigm. MaskFeat~\cite{wei2022masked} explored HOG features as reconstruction targets, finding structured features more effective than raw pixels for some tasks.

\textbf{CLIP-Guided Distillation in MIM.}
CLIP~\cite{radford2021learning} provides rich semantic features from vision-language pre-training that can serve as distillation targets for MIM. MILAN~\cite{hou2022milan} used CLIP attention maps for semantic-aware masking and caption guidance. BEiT v2~\cite{peng2022beit} trained a vector-quantized tokenizer using CLIP, producing more semantically meaningful visual tokens. MaskDistill~\cite{peng2022unified} provided a unified comparison of different teachers (CLIP, DINO, MAE) and reconstruction targets, establishing that CLIP-based patch-level distillation with smooth L1 loss yields strong results. BootMAE~\cite{dong2022bootstrapped} combined pixel reconstruction with a momentum encoder, while Data2Vec~\cite{baevski2022data2vec} generalized teacher-student distillation across speech, vision, and language. CMAE~\cite{huang2022contrastive} added contrastive objectives through a dual decoder with shifted view augmentation. SdAE~\cite{chen2022sdae} used layered masking with cosine similarity loss, achieving competitive results with fewer pre-training epochs.

\textbf{Masking Strategies.}
Beyond random and block masking, several works have explored more sophisticated strategies. AttMask~\cite{kakogeorgiou2022hide} used teacher attention to mask salient patches. SemMAE~\cite{li2022SemMAE} incorporated semantic segmentation to guide masking toward meaningful regions. Adversarial masking~\cite{shi2022adversarial} learned a masking subnet that identifies the most informative patches. Evolved Part Masking (EPM)~\cite{feng2023evolved} introduced adaptive masks that evolve during training, transitioning from grid-based to attention-guided patterns using an EM algorithm. Our work improves upon EPM by introducing hierarchical clustering with relative position bias for more coherent mask generation, while also demonstrating that in CLIP-distillation settings, these sophisticated strategies do not outperform simple block masking.

\textbf{Multi-Objective Learning in MIM.}
Most MIM methods optimize a single reconstruction target. Recent work has begun combining multiple objectives: BootMAE pairs pixel reconstruction with feature regression, while CMAE combines contrastive and reconstruction losses. However, the question of how to balance multiple objectives optimally remains underexplored. Standard multi-task learning methods such as uncertainty weighting~\cite{kendall2018multi} and GradNorm~\cite{chen2018gradnorm} apply global scalar weights per objective, but these cannot capture spatial variation in optimal loss emphasis across different image regions. Our work systematically investigates this multi-objective design space, revealing the extreme fragility of scalar weights and laying the groundwork for future adaptive weighting approaches.

\section{Method}
\label{sec:methodology}

\subsection{Overview}

Figure~\ref{fig:MEDiC} illustrates the MEDiC framework compared to three other mainstream MIM paradigms: (1)~raw-space reconstruction without distillation (MAE-style~\cite{he2022masked}), (2)~latent-space discriminative distillation via discrete tokens (BEiT-style~\cite{bao2021beit}), (3)~latent-space representative distillation via patch tokens from a teacher (MaskDistill-style~\cite{peng2022beit}), and (4)~our dual-space approach combining all three. By operating in both raw and latent spaces, MEDiC captures local spatial structure through pixel reconstruction and global semantics through teacher alignment.

MEDiC adopts a teacher-student architecture. The teacher is a frozen CLIP ViT-B/16 encoder~\cite{radford2021learning} that processes the full, unmasked image and provides semantic targets at both the patch and CLS token levels. The student is a ViT-Base encoder~\cite{dosovitskiy2020image} that observes only partially visible patches. Given an input image divided into $N$ patches, a binary mask $\boldsymbol{m} \in \{0,1\}^N$ partitions them into a visible set $\mathcal{V}$ and a masked set $\mathcal{M}$. A lightweight decoder reconstructs pixel values for the masked patches.

\subsection{Multi-Objective Distillation}
\label{sec:multi_objective}

The training objective combines three complementary loss terms:
\begin{equation}\label{eq:total_loss}
  \mathcal{L} = \lambda_{\text{rep}} \,\mathcal{L}_{\text{rep}}
              + \lambda_{\text{disc}} \,\mathcal{L}_{\text{disc}}
              + \lambda_{\text{pixel}} \,\mathcal{L}_{\text{pixel}}
\end{equation}

\noindent\textbf{Representative Distillation (Patch-Level).} We distill knowledge from the teacher at the patch level. The student outputs patch tokens $\hat{\boldsymbol{v}}_{s}^{\text{patch}}$ for the masked view, while the teacher outputs $\boldsymbol{v}_{t}^{\text{patch}}$ for the full view. The representative loss aligns these for masked positions:
\begin{equation}
\mathcal{L}_{\text{rep}}=\frac{1}{|\mathcal{M}|} \sum_{i \in \mathcal{M}} \operatorname{SmoothL1}\left(h\!\left(\hat{\boldsymbol{v}}_{s,i}^{\text{patch}}\right),\, \mathrm{LN}\left(\boldsymbol{v}_{t,i}^{\text{patch}}\right)\right),
\end{equation}
where $h(\cdot)$ is a linear projection from student to teacher dimension, $\operatorname{SmoothL1}$ combines L1 and L2 losses for robustness to outliers, and $\mathrm{LN}$ denotes layer normalization applied to teacher features. The same projection $h(\cdot)$ is shared with the CLS-level loss below.

\noindent\textbf{Discriminative Distillation (CLS-Level).} Both the teacher and student produce CLS tokens that capture global image-level information. We align these through a cross-entropy loss:
\begin{equation}
\mathcal{L}_{\text{disc}}=-P_{\boldsymbol{t}}^{[\mathrm{CLS}]}(\boldsymbol{v})^{\mathrm{T}} \log P_{\boldsymbol{s}}^{[\mathrm{CLS}]}(\hat{\boldsymbol{v}}),
\end{equation}
where $P_{\boldsymbol{t}}^{[\mathrm{CLS}]}(\boldsymbol{v}) = \mathrm{softmax}(\boldsymbol{v}_{t}^{\mathrm{CLS}})$ and $P_{\boldsymbol{s}}^{[\mathrm{CLS}]}(\hat{\boldsymbol{v}}) = \mathrm{softmax}(h(\hat{\boldsymbol{v}}_{s}^{\mathrm{CLS}}))$ are softmax distributions over the teacher and student CLS embeddings, with $h(\cdot)$ a linear projection from student to teacher dimension. This discriminative objective preserves global semantics while the representative loss captures local details.

\noindent\textbf{Pixel Reconstruction (Raw-Space).} Relying solely on latent-space distillation makes the student dependent on the teacher's representation capacity. To ground the learned features in raw visual content, we introduce a lightweight decoder $P_{\boldsymbol{d}}(\cdot)$ that reconstructs pixel values for masked patches:
\begin{equation}
\mathcal{L}_{\text{pixel}}=\frac{1}{|\mathcal{M}|} \sum_{i \in \mathcal{M}} \ell_2\left(\boldsymbol{x}_{i},\, \mathrm{LN}\left(\tilde{\boldsymbol{v}}_{d,i}\right)\right),
\end{equation}
where $\tilde{\boldsymbol{v}}_{d} = P_{\boldsymbol{d}}(\hat{\boldsymbol{v}}_{s}^{\text{patch}})$ is the decoder output and $\boldsymbol{x}_i$ are the original pixel values for patch $i$.

\subsection{Evolved Masking with Hierarchical Clustering}
\label{sec:evolved_masking}

\begin{figure}[H]
\begin{center}
\includegraphics[width=\linewidth]{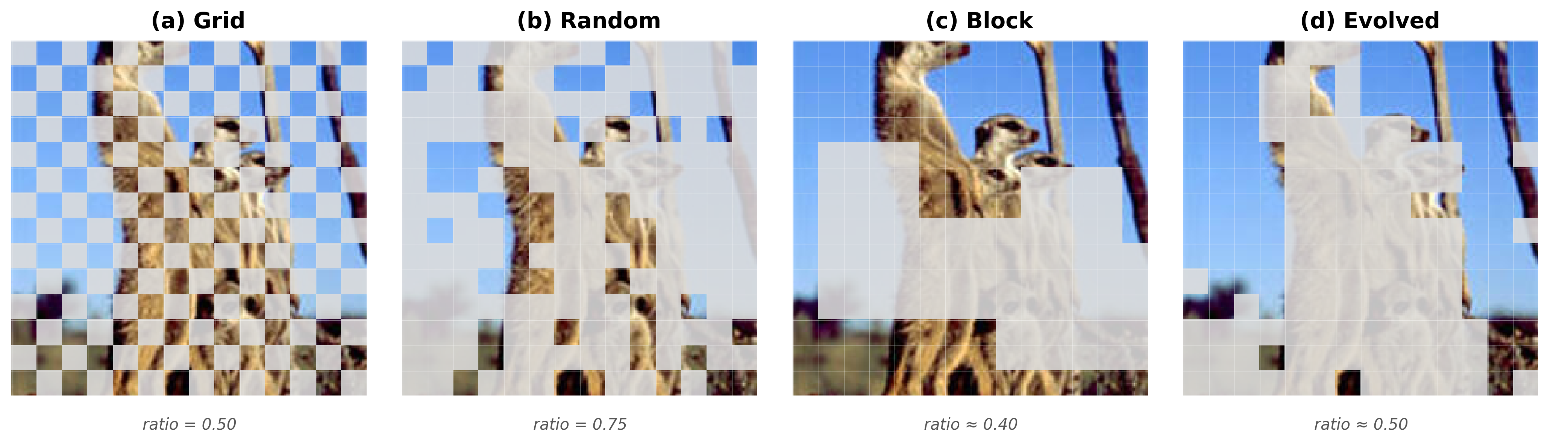}
\caption{Masking strategies in MIM. (a-c) Three standard approaches: grid, random, and block masking with different mask ratios. (d) Evolved masking uses attention-guided clustering to produce semantically coherent mask patterns that adapt during training.}
\label{fig:masking_strategies}
\end{center}
\end{figure}

The choice of which patches to mask affects what the model learns to reconstruct. While simple block masking is effective, evolved masking strategies~\cite{feng2023evolved} adapt the mask distribution during training based on the model's learned attention, potentially guiding the model toward more challenging reconstruction tasks.

We adopt the evolved masking framework of~\cite{feng2023evolved} with an improved clustering method. The masking strategy transitions from grid-based patterns to attention-guided selection over the course of training, controlled by:
\begin{equation}
    \alpha^{(k)} = \left( \frac{k}{K} \right)^{\gamma},
\end{equation}
where $k$ is the current epoch, $K$ is the total epochs, and $\gamma$ controls the transition rate.

At each epoch, attention maps $\mathbf{A} \in \mathbb{R}^{N \times N}$ from the last self-attention layer capture patch relationships. We improve upon the EM-based clustering of~\cite{feng2023evolved} by introducing hierarchical clustering (HC) with relative position bias. We compute a distance matrix:
\begin{equation}
      \mathbf{D}_{ij} = \zeta \cdot \left| \mathbf{A}_i - \mathbf{A}_j \right|^2 + (1 - \zeta) \cdot \mathbf{B}_{ij},
\end{equation}
where $\mathbf{A}_i, \mathbf{A}_j$ are attention vectors, $\mathbf{B}_{ij}$ is the relative position bias capturing spatial proximity, and $\zeta$ balances the two terms. Agglomerative clustering with average linkage groups patches into $C^{(k)}$ clusters:
\begin{equation}
    C^{(k)} = \left\lfloor C_{\text{min}} + (C_{\text{max}} - C_{\text{min}}) \cdot \alpha^{(k)} \right\rfloor,
\end{equation}
with the mask probability blending grid and cluster components:
\begin{equation}
    P_i^{(k)} = (1 - \alpha^{(k)}) \cdot P_i^{\text{grid}} + \alpha^{(k)} \cdot P_i^{\text{cluster}}.
\end{equation}

By incorporating relative position bias into the clustering, patches that are both semantically similar (based on attention) and spatially proximate are grouped together, producing more coherent mask patterns than EM-based approaches. We evaluate both EM and HC strategies in Section~\ref{sec:eval}.

\subsection{Encoding Strategies}
\label{sec:encoding}

The encoder can process patches in two modes that interact with the multi-objective framework:

\noindent\textbf{Dense encoding (BEiT-style).} All $N$ patch positions are processed, with masked positions replaced by a learnable mask token $\boldsymbol{e}_{\text{[MASK]}}$:
\begin{equation}
    \mathbf{x}_{\text{dense}} = \mathbf{x}_{\text{vis}} \odot (1 - \mathbf{m}) + \boldsymbol{e}_{\text{[MASK]}} \odot \mathbf{m}
\end{equation}

\noindent\textbf{Sparse encoding (MAE-style).} Only visible patches are processed, reducing computation by a factor of $(1-r)$ where $r$ is the mask ratio. Position embeddings are added before masking to preserve spatial information:
\begin{equation}
    \mathbf{x}_{\text{pos}} = \mathbf{x} + \mathbf{p}_{\text{emb}}, \quad
    \mathbf{x}_{\text{sparse}} = \mathbf{x}_{\text{pos}}[\neg\mathbf{m}]
\end{equation}

The encoding choice has implications for multi-objective learning. Sparse encoding is computationally efficient and yields stronger representations (Section~\ref{sec:encoding_results}), but produces encoder features only for visible patches, which constrains how losses on masked patches (pixel reconstruction) interact with encoder-level objectives (distillation). MEDiC uses sparse encoding by default.

\section{Experiments}
\label{sec:eval}

\subsection{Experimental Setup}

All experiments use ImageNet-1K~\cite{deng2009imagenet} with a ViT-Base/16 student encoder and a frozen CLIP ViT-B/16 teacher. We pre-train for 300 epochs with block masking at 40\% ratio, batch size 2048, AdamW optimizer ($\beta_1\!=\!0.9$, $\beta_2\!=\!0.999$), weight decay 0.05, peak learning rate $1.5 \times 10^{-3}$ with cosine schedule and 10-epoch warmup. The decoder has 8 transformer layers (512 hidden, 16 heads). We adopt MaskDistill~\cite{peng2022unified} as our primary baseline given its use of the same CLIP teacher. Full hyperparameters are in the Appendix.

We evaluate across four protocols: (1)~\textbf{kNN} ($k\!=\!20$, cosine similarity) on frozen representations, (2)~\textbf{linear probing} following BEiT~\cite{bao2021beit} protocol, (3)~\textbf{ImageNet fine-tuning} with layer-wise LR decay, and (4)~\textbf{ADE20K segmentation} with an FCN head (linear, frozen backbone).

\subsection{Comparison with State-of-the-Art}
\label{sec:sota_comparison}

We assess learned representations on three datasets of increasing difficulty: \textbf{Imagenette}~\cite{imagewang} (easily distinguished ImageNet classes), \textbf{Imagewoof}~\cite{imagewang} (fine-grained dog breeds), and the full \textbf{ImageNet-1K} test set.

\begin{table*}[t!]
  \centering
  \caption{kNN classification accuracy using frozen representations. We report Top-1 and Top-5 accuracies on Imagenette, Imagewoof, and ImageNet-1K.}\label{tab:knn_frozen_repr}
  \vspace{2mm}
  \setlength{\tabcolsep}{3.5mm}{
  \begin{tabular}{l|c|c|cc|cc|c}
  \toprule
  \multicolumn{1}{c}{}&\multicolumn{1}{c}{}&\multicolumn{1}{c|}{}&\multicolumn{2}{c|}{Imagenette}&\multicolumn{2}{c|}{Imagewoof}&\multicolumn{1}{c}{ImageNet-1K}\\
	\hline
	{} & Backbone & Epochs & top-1 & top-5 & top-1 & top-5 & top-1 \\
	\hline
	MAE~\cite{he2022masked} & ViT-B & 800 & 54.75 & 89.35 & 30.29 & 76.79 & 9.53 \\
	BEiT~\cite{bao2021beit}  & ViT-B & 800 & 30.06 & 75.69 & 20.26 & 70.40 & 1.93 \\
    BootMAE~\cite{dong2022bootstrapped}  & ViT-B & 800 & 70.70 & 94.22 & 41.74 & 84.53 & 19.61 \\
    CMAE~\cite{huang2022contrastive}  & ViT-B & 800 & 77.91 & 96.33 & 42.84 & 85.24 & 21.17 \\
	SimMIM~\cite{xie2022simmim} & ViT-B & 400 & 51.44 & 87.21 & 30.16 & 74.65 & 8.40 \\
    SemMAE~\cite{li2022SemMAE} & ViT-B & 800 & 70.04 & 93.55 & 44.49 & 85.87 & 19.13 \\
	MaskDistill~\cite{peng2022unified} & ViT-B & 300 & 76.00 & 95.77 & 49.33 & 89.31 & 68.59\\
	\midrule
	\textbf{MEDiC (Ours)} & ViT-B & 300 & \textbf{85.96} &  \textbf{98.47} & \textbf{63.45} & \textbf{95.93} & \textbf{73.92} \\
	\bottomrule
	\end{tabular}}
\end{table*}

Table~\ref{tab:knn_frozen_repr} shows that MEDiC outperforms all compared methods in kNN accuracy with a frozen encoder. The improvement over MaskDistill is particularly notable: +9.96 points on Imagenette, +14.12 on Imagewoof, and +5.33 on ImageNet-1K. The larger gains on Imagewoof (fine-grained dog breeds) suggest that multi-level distillation enhances the balance between local and global features, which is especially valuable for distinguishing visually similar categories.

Compared to methods without a CLIP teacher (MAE, BEiT, BootMAE, CMAE, SimMIM, SemMAE), MEDiC achieves substantially higher kNN accuracy despite using only 300 pre-training epochs versus their 400-800, demonstrating the effectiveness of CLIP-based multi-objective distillation.

\begin{figure}[H]
\begin{center}
\includegraphics[width=\linewidth]{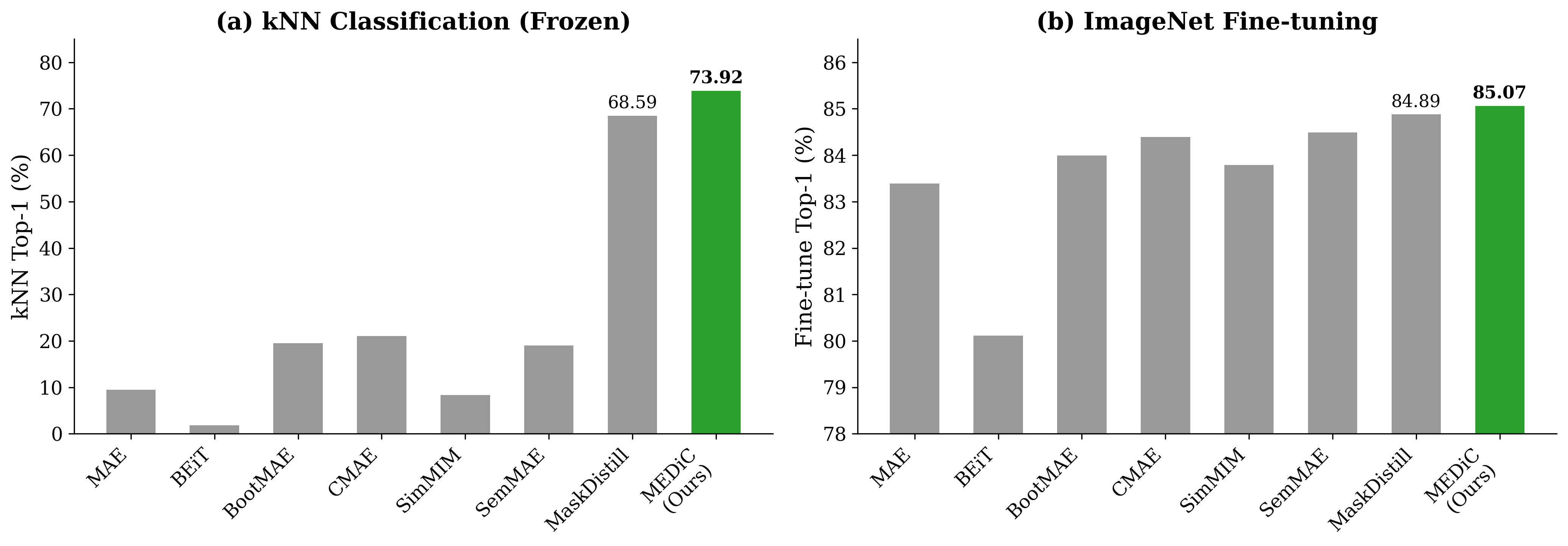}
\caption{MEDiC achieves strong kNN and fine-tuning performance through multi-objective distillation from CLIP, outperforming methods that operate in either raw or latent space alone.}\label{fig:shoutout}
\end{center}
\end{figure}

\begin{table}[H]
    \centering
    \caption{Combined evaluation on ImageNet-1K: kNN (frozen), linear probing, and fine-tuning accuracy. $\dagger$: publicly available checkpoints. $\ddagger$: our reproduction.}
    \label{tab:combined_eval}
    \resizebox{\linewidth}{!}{%
        \begin{tabular}{lccccc}
            \toprule
            \textbf{Method} & \textbf{Backbone} & \textbf{Epochs} & \textbf{kNN} & \textbf{Linear} & \textbf{Fine-tune} \\
            \midrule
            MAE$\dagger$ & ViT-L & 400 & 9.53 & 54.60 & 83.4 \\
            BEiT$\dagger$ & ViT-B & 800 & 1.93 & 52.10 & 80.12 \\
            BootMAE$\dagger$ & ViT-B & 800 & 19.61 & 52.07 & 84.0 \\
            CMAE$\dagger$ & ViT-B & 800 & 21.17 & 60.00 & 84.4 \\
            SimMIM$\dagger$ & ViT-B & 400 & 8.40 & 58.09 & 83.8 \\
            SemMAE$\dagger$ & ViT-B & 800 & 19.13 & 54.62 & 84.5 \\
            MaskDistill$\ddagger$ & ViT-B & 300 & 68.59 & 54.07 & 84.89 \\
            \midrule
            \textbf{MEDiC (Ours)} & ViT-B & 300 & \textbf{73.92} & \textbf{60.50} & \textbf{85.07} \\
            \bottomrule
        \end{tabular}
    }
\end{table}

Table~\ref{tab:combined_eval} consolidates results across three evaluation protocols. MEDiC achieves 60.50\% linear probe accuracy, outperforming all compared methods including CMAE (60.00\%). Fine-tuning accuracy reaches 85.07\%, a modest but consistent improvement over MaskDistill (84.89\%), demonstrating that the multi-objective framework produces representations that adapt well to end-to-end training.

\begin{table}[H]
  \centering
  \caption{ADE20K semantic segmentation (mIoU \%) with UperNet decoder and end-to-end fine-tuning (160K iterations). $\dagger$: published results. $\ddagger$: our reproduction at 300 epochs.}\label{tab:semseg}
  \vspace{2mm}
  \begin{tabular}{lccc}
	\toprule
	Method & Backbone & Epochs & mIoU \\
	\midrule
	BEiT$\dagger$~\cite{bao2021beit} & ViT-B & 800 & 45.6\\
	MAE$\dagger$~\cite{he2022masked} & ViT-B & 1600 & 48.1\\
	BEiT v2$\dagger$~\cite{peng2022beit} & ViT-B & 300 & 52.7\\
	MILAN$\dagger$~\cite{hou2022milan} & ViT-B & 400 & 52.7\\
	CAE v2$\dagger$~\cite{zhang2023caev2} & ViT-B & 300 & 53.4\\
	MaskDistill$\ddagger$~\cite{peng2022unified} & ViT-B & 300 & 53.8\\
	\midrule
	\textbf{MEDiC (Ours)} & ViT-B & 300 & 52.5 \\
	\bottomrule
	\end{tabular}
\end{table}

Table~\ref{tab:semseg} presents semantic segmentation on ADE20K using UperNet with end-to-end fine-tuning. MEDiC achieves 52.5\% mIoU, competitive with BEiT v2 and MILAN (52.7\%) at comparable or fewer pre-training epochs. The gap to MaskDistill (53.8\%) and CAE v2 (53.4\%) suggests room for improvement in how multi-objective features transfer to dense prediction tasks.

\subsection{Loss Component Ablation}
\label{sec:ablation}

\begin{table}[H]
  \centering
  \caption{Effect of each loss component. Combining all three objectives yields the strongest kNN accuracy.}\label{tab:compare_losses}
  \vspace{2mm}
  \begin{tabular}{ccc|c}
  \hline
 Latent-Rep & Raw-Pix & Latent-Disc & kNN Top-1 \\
 \hline
 $\circ$ & $\circ$ & $\checkmark$ & 14.1 \\
 $\circ$ & $\checkmark$ & $\circ$ & 9.5 \\
 $\checkmark$ & $\circ$ & $\circ$ & 68.6\\
 $\checkmark$ & $\checkmark$ & $\circ$ & 71.4\\
 $\checkmark$ & $\circ$ & $\checkmark$ & 72.3 \\
 $\checkmark$ & $\checkmark$ & $\checkmark$ & \textbf{73.9}\\
 \hline
  \end{tabular}
\end{table}

Table~\ref{tab:compare_losses} isolates the contribution of each loss component. Patch-level distillation alone achieves 68.6\%, confirming that CLIP features provide a strong learning signal. Adding pixel reconstruction raises this to 71.4\% (+2.8), while adding CLS alignment instead yields 72.3\% (+3.7). The full combination reaches 73.9\%, demonstrating that all three objectives are complementary. Note that pixel reconstruction alone (9.5\%) matches MAE's kNN performance, as expected since this configuration is equivalent to a pixel-only masked autoencoder.

\subsection{Loss Weight Sensitivity}
\label{sec:weight_sensitivity}

\begin{figure}[H]
\begin{center}
\includegraphics[width=\linewidth]{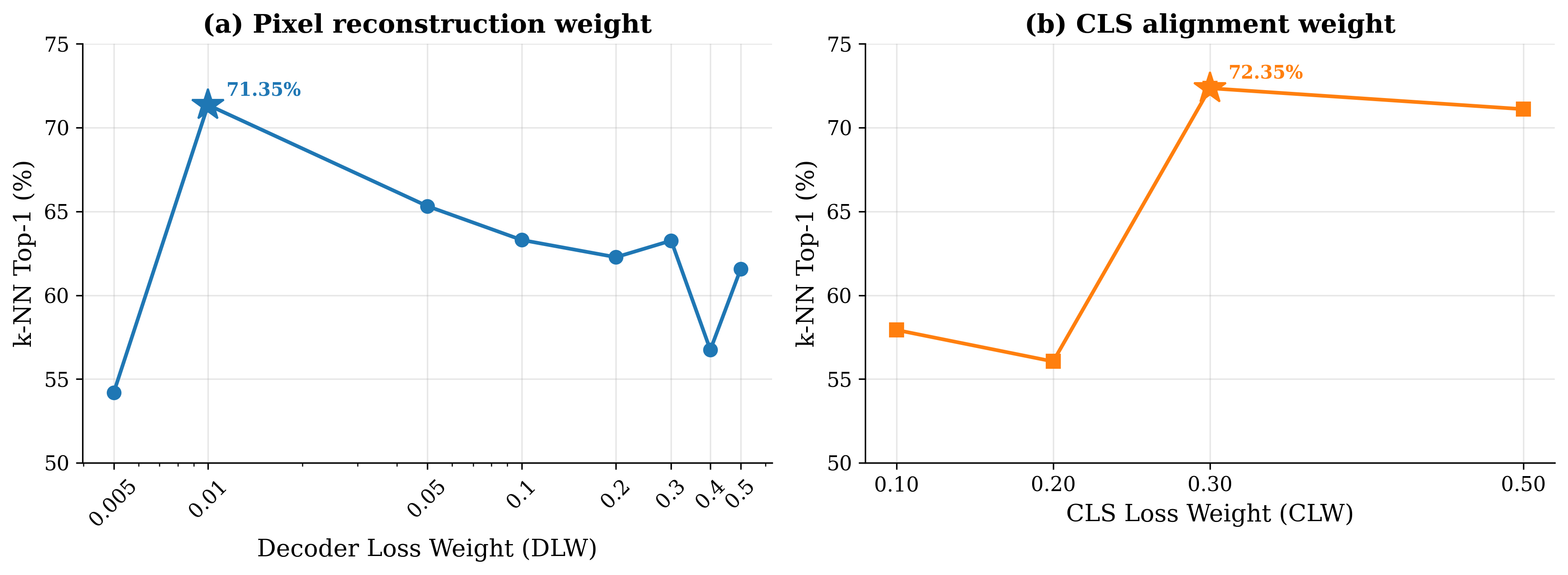}
\caption{Loss weight sensitivity. (a) Pixel reconstruction weight (DLW) has a sharp optimum at 0.01; higher values degrade kNN by up to 17 points. (b) CLS alignment weight (CLW) peaks at 0.30 with a sudden drop at 0.20. Both curves reveal the fragility of global scalar weights. The combined optimum (DLW=0.01, CLW=0.30) yields 73.92\% kNN and 85.07\% fine-tuning. Full sweep data in the Appendix.}
\label{fig:weight_sweep}
\end{center}
\end{figure}

Figure~\ref{fig:weight_sweep} reveals extreme sensitivity in the loss weight landscape. We denote the pixel reconstruction weight as DLW ($= \lambda_{\text{pixel}}$) and the CLS alignment weight as CLW ($= \lambda_{\text{disc}}$), with the patch distillation weight $\lambda_{\text{rep}}$ fixed at 1.0. The pixel reconstruction weight (DLW) has a sharp optimum at 0.01: moving to 0.50 drops kNN accuracy by nearly 10 points (71.35\% $\to$ 61.57\%), while a small perturbation to 0.005 causes a catastrophic drop to 54.19\%. The CLS weight (CLW) shows a similarly narrow effective range, with CLW=0.30 achieving 72.33\% while CLW=0.20 drops to 56.05\%. The combined optimum of DLW=0.01 and CLW=0.30 yields 73.92\% kNN and 85.07\% fine-tuning accuracy.

This fragility is a fundamental limitation of global scalar weighting: a single weight applies uniformly to all patches regardless of their content. Different image regions may benefit from different loss emphasis, a direction we identify for future investigation.

\subsection{Masking Strategy Analysis}
\label{sec:masking_analysis}

\begin{figure*}[t!]
    \centering
    \includegraphics[width=\textwidth]{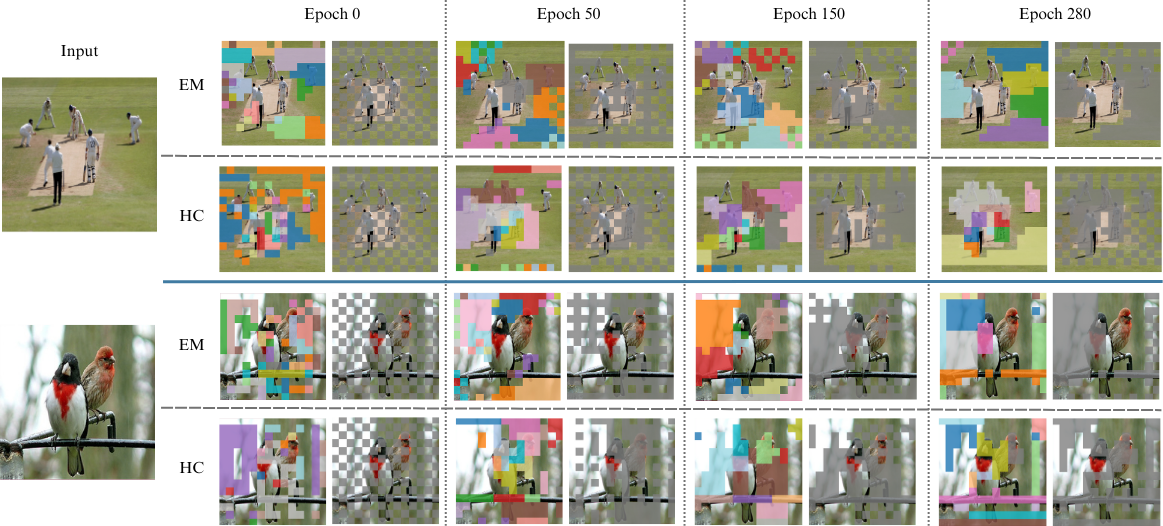}
    \caption{Evolved masking across training epochs. For each input image, the top row shows EM-based masks and the bottom row shows HC-based masks. HC produces more spatially coherent groupings that align with semantic content.}
    \label{fig:epoch_masks}
\end{figure*}

\begin{table}[H]
    \centering
    \caption{Masking strategy comparison. Despite HC producing more coherent masks (Fig.~\ref{fig:epoch_masks}), block masking achieves the highest kNN accuracy.}
    \label{tab:masking_summary}
    \vspace{2mm}
    \begin{tabular}{l|c}
    \hline
     Masking Method  & kNN Top-1 \\
    \hline
     Block Masking & \textbf{68.59} \\
     Evolved Masking + EM & 47.52 \\
     Evolved Masking + HC (Ours) & 64.56 \\
    \hline
    \end{tabular}
\end{table}

\begin{figure}[H]
\begin{center}
\includegraphics[width=\linewidth]{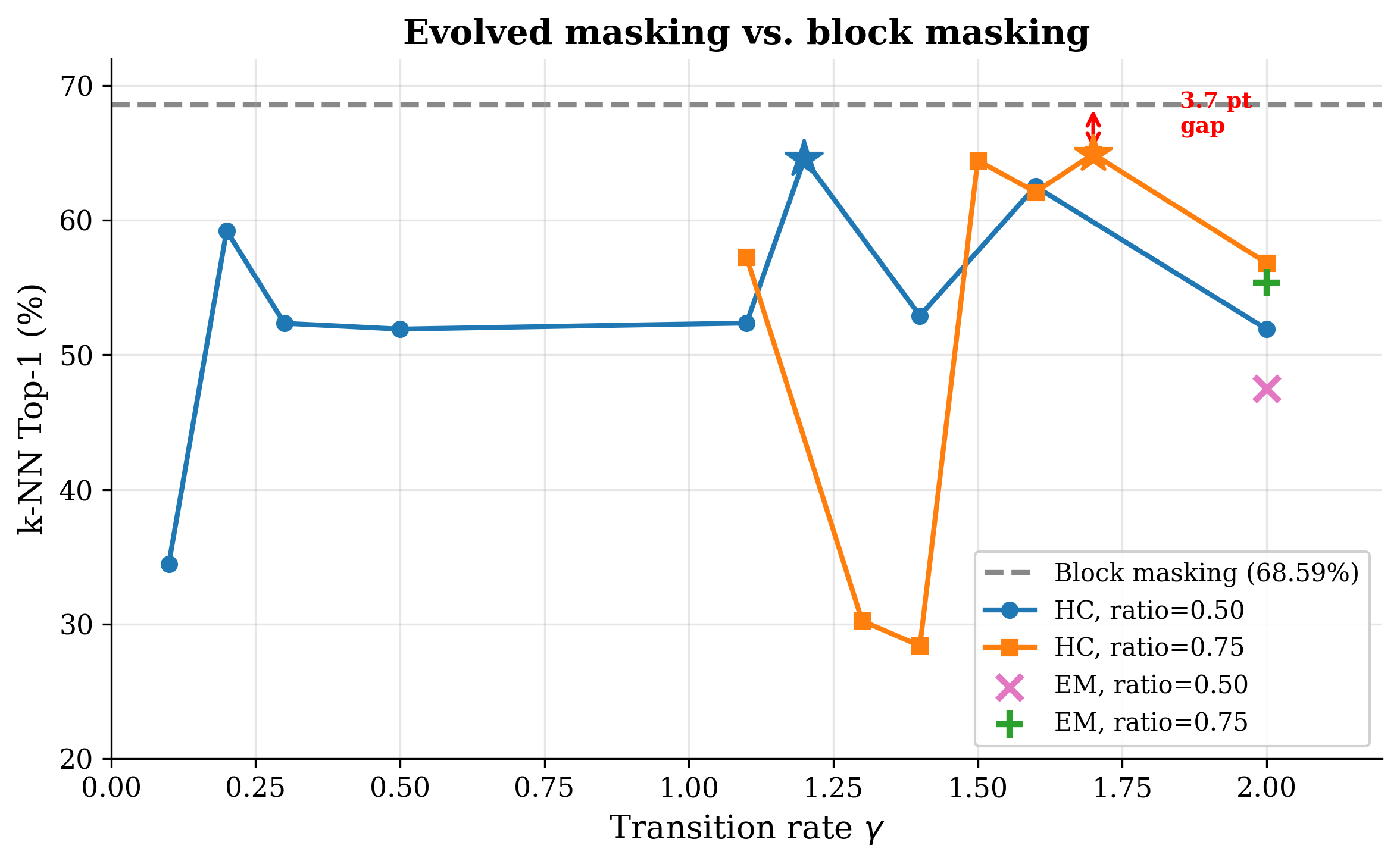}
\caption{Evolved masking vs. block masking. HC masking is swept across transition rates at two mask ratios. The block masking baseline (68.59\%, dashed line) outperforms all evolved configurations. The best HC result (64.93\%) leaves a 3.7-point gap.}
\label{fig:masking_sweep}
\end{center}
\end{figure}

Table~\ref{tab:masking_summary} compares the three masking strategies at their best configurations. HC-based evolved masking ($\gamma\!=\!1.2$, $\zeta\!=\!0.5$, ratio=0.5) substantially improves over EM (64.56\% vs 47.52\%), demonstrating the value of incorporating relative position bias into the clustering. However, block masking at 68.59\% still outperforms both evolved approaches by at least 4 points.

A comprehensive sweep across 26 evolved masking configurations (see Appendix) confirms this finding across different gamma values, zeta values, and mask ratios. No evolved configuration closes the gap with block masking. We attribute this to the CLIP teacher's inherent semantic awareness: since the teacher's representations already encode spatial and semantic structure, attention-guided masking provides redundant information that the simpler block strategy avoids.

\subsection{Dense vs.\ Sparse Encoding}
\label{sec:encoding_results}

\begin{table}[H]
  \centering
  \caption{Dense vs.\ sparse encoding comparison (kNN@20). Sparse encoding consistently outperforms dense across objective combinations.}
  \label{tab:encoding}
  \begin{tabular}{lcc}
    \toprule
    Objectives & Sparse & Dense \\
    \midrule
    Token + CLS & \textbf{75.7} & 74.1 \\
    Token + Pixel & \textbf{71.5} & 66.9 \\
    Token + Pixel + CLS & \textbf{73.9} & 69.3 \\
    \bottomrule
  \end{tabular}
\end{table}

Table~\ref{tab:encoding} compares dense and sparse encoding for different objective combinations. Sparse encoding (processing only visible patches) consistently outperforms dense encoding (processing all patches with mask tokens) by 1.6 to 4.6 kNN points. The gap is largest for Token+Pixel objectives (+4.6), suggesting that dense encoding's mask tokens may introduce noise that interferes with pixel reconstruction. Sparse encoding also reduces computation by a factor of $(1-r)$, making it preferable on both accuracy and efficiency grounds.

\section{Conclusion}
\label{sec:conclusion}

We presented MEDiC, a multi-objective masked distillation framework that combines pixel reconstruction with patch-level and CLS-level CLIP distillation in a dual-space setting. Our systematic investigation yielded several findings relevant to the design of multi-objective MIM systems. All three learning objectives provide complementary information, with the full combination outperforming any subset. Hierarchical clustering with relative position bias produces more semantically coherent masks than EM-based evolved masking, though block masking remains superior in teacher-guided distillation settings where the CLIP teacher already provides semantic awareness. The optimal scalar loss weights are extremely fragile, with small perturbations causing drops of up to 17 percentage points in kNN accuracy, suggesting a fundamental limitation of global weighting strategies for spatially heterogeneous objectives. Sparse encoding consistently outperforms dense encoding by 1.6 to 4.6 kNN points across objective combinations, with the largest gap occurring when pixel reconstruction is included.

MEDiC achieves 73.9\% kNN and 85.1\% fine-tuning accuracy on ImageNet-1K with ViT-Base at 300 pre-training epochs. The weight sensitivity analysis points to per-patch adaptive loss weighting as a promising direction for addressing the spatial heterogeneity in multi-objective MIM, where different image regions may benefit from different loss emphasis.

{
    \small
    \bibliographystyle{ieeenat_fullname}
    \bibliography{main}

@article{devlin2019bert,
  title     = {BERT: Pre-training of Deep Bidirectional Transformers for Language Understanding},
  author    = {Jacob Devlin and Ming-Wei Chang and Kenton Lee and Kristina Toutanova},
  journal   = {North American Chapter of the Association for Computational Linguistics},
  year      = {2019},
  doi       = {10.18653/v1/N19-1423},
  bibSource = {Semantic Scholar https://www.semanticscholar.org/paper/df2b0e26d0599ce3e70df8a9da02e51594e0e992}
}

@article{dosovitskiy2020image,
  title={An image is worth 16x16 words: Transformers for image recognition at scale},
  author={Dosovitskiy, Alexey and Beyer, Lucas and Kolesnikov, Alexander and Weissenborn, Dirk and Zhai, Xiaohua and Unterthiner, Thomas and Dehghani, Mostafa and Minderer, Matthias and Heigold, Georg and Gelly, Sylvain and others},
  journal={arXiv preprint arXiv:2010.11929},
  year={2020}
}

@article{bao2021beit,
  title={Beit: Bert pre-training of image transformers},
  author={Bao, Hangbo and Dong, Li and Piao, Songhao and Wei, Furu},
  journal={arXiv preprint arXiv:2106.08254},
  year={2021}
}

@inproceedings{he2022masked,
  title={Masked autoencoders are scalable vision learners},
  author={He, Kaiming and Chen, Xinlei and Xie, Saining and Li, Yanghao and Doll{\'a}r, Piotr and Girshick, Ross},
  booktitle={Proceedings of the IEEE/CVF conference on computer vision and pattern recognition},
  pages={16000--16009},
  year={2022}
}

@article{zhou2021ibot,
  title={ibot: Image bert pre-training with online tokenizer},
  author={Zhou, Jinghao and Wei, Chen and Wang, Huiyu and Shen, Wei and Xie, Cihang and Yuille, Alan and Kong, Tao},
  journal={arXiv preprint arXiv:2111.07832},
  year={2021}
}

@inproceedings{wei2022masked,
  title={Masked feature prediction for self-supervised visual pre-training},
  author={Wei, Chen and Fan, Haoqi and Xie, Saining and Wu, Chao-Yuan and Yuille, Alan and Feichtenhofer, Christoph},
  booktitle={Proceedings of the IEEE/CVF Conference on Computer Vision and Pattern Recognition},
  pages={14668--14678},
  year={2022}
}

@inproceedings{baevski2022data2vec,
  title={Data2vec: A general framework for self-supervised learning in speech, vision and language},
  author={Baevski, Alexei and Hsu, Wei-Ning and Xu, Qiantong and Babu, Arun and Gu, Jiatao and Auli, Michael},
  booktitle={International Conference on Machine Learning},
  pages={1298--1312},
  year={2022},
  organization={PMLR}
}

@inproceedings{kakogeorgiou2022hide,
  title={What to hide from your students: Attention-guided masked image modeling},
  author={Kakogeorgiou, Ioannis and Gidaris, Spyros and Psomas, Bill and Avrithis, Yannis and Bursuc, Andrei and Karantzalos, Konstantinos and Komodakis, Nikos},
  booktitle={European Conference on Computer Vision},
  pages={300--318},
  year={2022},
  organization={Springer}
}

@article{huang2022contrastive,
  title={Contrastive masked autoencoders are stronger vision learners},
  author={Huang, Zhicheng and Jin, Xiaojie and Lu, Chengze and Hou, Qibin and Cheng, Ming-Ming and Fu, Dongmei and Shen, Xiaohui and Feng, Jiashi},
  journal={arXiv preprint arXiv:2207.13532},
  year={2022}
}

@inproceedings{chen2022sdae,
  title={Sdae: Self-distillated masked autoencoder},
  author={Chen, Yabo and Liu, Yuchen and Jiang, Dongsheng and Zhang, Xiaopeng and Dai, Wenrui and Xiong, Hongkai and Tian, Qi},
  booktitle={European Conference on Computer Vision},
  pages={108--124},
  year={2022},
  organization={Springer}
}

@article{hou2022milan,
  title={Milan: Masked image pretraining on language assisted representation},
  author={Hou, Zejiang and Sun, Fei and Chen, Yen-Kuang and Xie, Yuan and Kung, Sun-Yuan},
  journal={arXiv preprint arXiv:2208.06049},
  year={2022}
}

@article{peng2022beit,
  title={Beit v2: Masked image modeling with vector-quantized visual tokenizers},
  author={Peng, Zhiliang and Dong, Li and Bao, Hangbo and Ye, Qixiang and Wei, Furu},
  journal={arXiv preprint arXiv:2208.06366},
  year={2022}
}

@article{radford2021learning,
  title     = {Learning Transferable Visual Models From Natural Language Supervision},
  author    = {Alec Radford and Jong Wook Kim and Chris Hallacy and A. Ramesh and Gabriel Goh and Sandhini Agarwal and Girish Sastry and Amanda Askell and Pamela Mishkin and Jack Clark and Gretchen Krueger and Ilya Sutskever},
  journal   = {International Conference on Machine Learning},
  year      = {2021},
  bibSource = {Semantic Scholar https://www.semanticscholar.org/paper/6f870f7f02a8c59c3e23f407f3ef00dd1dcf8fc4}
}

@article{peng2022unified,
  title={A unified view of masked image modeling},
  author={Peng, Zhiliang and Dong, Li and Bao, Hangbo and Ye, Qixiang and Wei, Furu},
  journal={arXiv preprint arXiv:2210.10615},
  year={2022}
}

@inproceedings{shi2022adversarial,
  title={Adversarial masking for self-supervised learning},
  author={Shi, Yuge and Siddharth, N and Torr, Philip and Kosiorek, Adam R},
  booktitle={International Conference on Machine Learning},
  pages={20026--20040},
  year={2022},
  organization={PMLR}
}

@inproceedings{feng2023evolved,
  title={Evolved Part Masking for Self-Supervised Learning},
  author={Feng, Zhanzhou and Zhang, Shiliang},
  booktitle={Proceedings of the IEEE/CVF Conference on Computer Vision and Pattern Recognition},
  pages={10386--10395},
  year={2023}
}

@article{zhang2023caev2,
  title={CAE v2: Context Autoencoder with CLIP Target},
  author={Zhang, Xinyu and Wu, Jiahui and Peng, Zhiliang and Liu, Hao and Dong, Li and De Mello, Sanja and Zeng, Zhifeng and Aneja, Jianfeng and Zhu, Jiaqi and Yan, Shuicheng and Wei, Furu},
  journal={Transactions on Machine Learning Research},
  year={2023}
}

@misc{imagewang,
    author    = "Jeremy Howard",
    title     = "Imagewang",
    year      = {2019},
    url       = "https://github.com/fastai/imagenette/"
}

@inproceedings{dong2022bootstrapped,
  title={Bootstrapped masked autoencoders for vision BERT pretraining},
  author={Dong, Xiaoyi and Bao, Jianmin and Zhang, Ting and Chen, Dongdong and Zhang, Weiming and Yuan, Lu and Chen, Dong and Wen, Fang and Yu, Nenghai},
  booktitle={European Conference on Computer Vision},
  pages={247--264},
  year={2022},
  organization={Springer}
}

@inproceedings{xie2022simmim,
  title={Simmim: A simple framework for masked image modeling},
  author={Xie, Zhenda and Zhang, Zheng and Cao, Yue and Lin, Yutong and Bao, Jianmin and Yao, Zhuliang and Dai, Qi and Hu, Han},
  booktitle={Proceedings of the IEEE/CVF Conference on Computer Vision and Pattern Recognition},
  pages={9653--9663},
  year={2022}
}

@inproceedings{caron2021emerging,
  title={Emerging properties in self-supervised vision transformers},
  author={Caron, Mathilde and Touvron, Hugo and Misra, Ishan and J{\'e}gou, Herv{\'e} and Mairal, Julien and Bojanowski, Piotr and Joulin, Armand},
  booktitle={Proceedings of the IEEE/CVF International Conference on Computer Vision},
  pages={9650--9660},
  year={2021}
}

@inproceedings{chen2018gradnorm,
  title={GradNorm: Gradient normalization for adaptive loss balancing in deep multitask networks},
  author={Chen, Zhao and Badrinarayanan, Vijay and Lee, Chen-Yu and Rabinovich, Andrew},
  booktitle={International Conference on Machine Learning},
  pages={794--803},
  year={2018}
}

@inproceedings{deng2009imagenet,
  title={ImageNet: A large-scale hierarchical image database},
  author={Deng, Jia and Dong, Wei and Socher, Richard and Li, Li-Jia and Li, Kai and Fei-Fei, Li},
  booktitle={IEEE Conference on Computer Vision and Pattern Recognition},
  pages={248--255},
  year={2009}
}

@inproceedings{kendall2018multi,
  title={Multi-task learning using uncertainty to weigh losses for scene geometry and semantics},
  author={Kendall, Alex and Gal, Yarin and Cipolla, Roberto},
  booktitle={Proceedings of the IEEE Conference on Computer Vision and Pattern Recognition},
  pages={7482--7491},
  year={2018}
}

@inproceedings{li2022SemMAE,
  title={SemMAE: Semantic-guided masking for learning masked autoencoders},
  author={Li, Gang and Zheng, Heliang and Liu, Daqing and Wang, Chaoyue and Su, Bing and Zheng, Changwen},
  booktitle={Advances in Neural Information Processing Systems},
  volume={35},
  year={2022}
}
}

\newpage
\appendix
\clearpage
\section{Overview and Hierarchical Clustering}

This supplement provides additional details and results: (B)~our hierarchical clustering algorithm with pseudocode, (C)~training hyperparameters, (D)~masking hyperparameters and evolved mask examples, (E)~full loss weight sweep data, and (F)~full evolved masking sweep across 26 configurations.

\noindent\textbf{Hierarchical Clustering Algorithm.}
We use average-linkage hierarchical clustering to group patches according to both attention map similarity and relative position. Each batch of images has an attention matrix $\mathbf{A}^b$ and a relative position bias matrix $\mathbf{B}$. We combine them with a weighting factor $\zeta$ to control how strongly position influences clustering. Algorithm~\ref{alg:hierarchical_clustering} provides the full pseudocode.

\begin{algorithm}[H]
    \caption{Hierarchical Clustering with Relative Position}
    \label{alg:hierarchical_clustering}
    \begin{algorithmic}
        \STATE {\bfseries Input:} $\mathbf{A} \in \mathbb{R}^{B \times N \times N}$ (attention maps from the student encoder),
        $\mathbf{B} \in \mathbb{R}^{N \times N}$ (relative position bias matrix),
        $C^{(k)}$ (number of clusters),
        $\zeta \in [0, 1]$ (weighting factor)
        \STATE {\bfseries Output:} $\mathbf{c}^b \in \{1, \dots, C^{(k)}\}$ for each image $b$
        \FOR{$b = 1$ {\bfseries to} $B$}
            \STATE $\mathbf{A}^b \leftarrow \mathbf{A}[b]$
            \COMMENT{Extract attention map for the $b$th image}
            \STATE $\mathbf{D}_{\text{attn}} \leftarrow
            \bigl[\|\mathbf{A}^b_i - \mathbf{A}^b_j\|^2\bigr]_{i,j=1}^N$
            \COMMENT{Compute attention difference matrix}
            \STATE $\mathbf{D} \leftarrow
            \zeta \cdot \mathbf{D}_{\text{attn}}
            + (1 - \zeta) \cdot \mathbf{B}$
            \COMMENT{Combine with relative position bias}
            \STATE $\mathbf{D} \leftarrow
            \tfrac{1}{2}\bigl(\mathbf{D} + \mathbf{D}^\top\bigr),
            \quad \mathbf{D}_{ii} = 0 \; \forall i$
            \COMMENT{Symmetrize and set diagonal to zero}
            \STATE Convert $\mathbf{D}$ to condensed form for hierarchical clustering
            \STATE Perform hierarchical clustering (average linkage) on $\mathbf{D}$ to obtain $\mathbf{c}^b$
        \ENDFOR
        \STATE {\bfseries Return:} $\{\mathbf{c}^b\}_{b=1}^B$
    \end{algorithmic}
\end{algorithm}

\section{Training Details}
\label{app:hyperparameters}

\subsection{Pre-training}
\label{app:pretraining}
\begin{table}[H]
\centering
\caption{Hyperparameters for pre-training on ImageNet-1K using ViT-Base model.}
\label{tbl:pretrain:hyperparams}
\small
\scalebox{0.98}{
    \begin{tabular}{l|c}
    \toprule
    \bf Hyperparameters & \bf Base Size \\
    \midrule
    Layers & 12 \\
    Hidden size & 768 \\
    FFN inner hidden size & 3072 \\
    Attention heads & 12 \\
    Layer scale & 0.1 \\
    Patch size & $16 \times 16$ \\
    Shared relative positional embeddings & \cmark \\
    \midrule
    Training epochs & 300 \\
    Batch size & 2048 \\
    Adam $\epsilon$ & 1e-8 \\
    Adam $\beta$ & (0.9, 0.999) \\
    Peak learning rate & 1.5e-3 \\
    Minimal learning rate & 1e-5 \\
    Learning rate schedule & Cosine \\
    Warmup epochs & 10 \\
    \midrule
    Stoch. depth & 0.1 \\
    Gradient clipping & 3.0 \\
    Dropout & \xmark \\
    Weight decay & 0.05 \\
    \midrule
    Data Augment & RandomResizeAndCrop \\
    Input resolution & $224 \times 224$ \\
    Color jitter & 0.4 \\
    \midrule
    Decoder Layers & 8 \\
    Decoder Hidden size & 512 \\
    Decoder FFN inner hidden size & 2048 \\
    Decoder Attention heads & 16 \\
    Decoder Loss Norm & \xmark \\
    Decoder Loss & L2 \\
    CLS Loss & Cross Entropy \\
    Patch Loss & Smooth L1 \\

    \bottomrule
    \end{tabular}
}
\end{table}

\subsection{Linear Probing}
\label{app:linearprobing}
For our linear probing experiments, we utilized the BEiT framework to assess the quality of representations learned by our model, which was pre-trained for 300 epochs. Our pre-trained model was directly integrated into the BEiT linear probing setup. To ensure consistency, we also evaluated the official pre-trained weights of other models using the same configuration.

We adopted the BEiT-base architecture with a patch size of 16 and an input resolution of 224×224 for the linear probing implementation. Consistent with the original BEiT settings, we maintained configurations such as relative positional embeddings and layer scale initialization. Following standard linear evaluation protocols, a supervised linear classifier was appended to the frozen backbone. The training was conducted using the AdamW optimizer with a peak learning rate of $5 \times 10^{-4}$, and the models were trained for 100 epochs on the ImageNet-1K dataset. Linear probing hyperparameter setups are shown in Table~\ref{tbl:lp:imagenet:hyperparams}. 

\begin{table}[H]
\centering
\caption{Hyperparameters for linear-probing on ImageNet-1K.}
\label{tbl:lp:imagenet:hyperparams}
\small
\scalebox{0.95}{
    \begin{tabular}{l|c}
    \toprule
    \bf Hyperparameters & \bf ViT-B/16 \\
    \midrule
    Peak learning rate & 5e-4\\
    Epochs & 100\\
    Warmup epochs & 20\\
    Batch size & 1024 \\
    Adam $\epsilon$ & 1e-8  \\
    Adam $\beta$ & (0.9, 0.999) \\
    Minimal learning rate & 1e-6 \\
    Learning rate schedule & Cosine \\
    \midrule
    Stoch. depth & 0.1 \\
    Repeated Aug & \xmark \\
    Weight decay & 0.05 \\
    Dropout & \xmark \\
    Gradient clipping & \xmark \\
    \midrule
    Input resolution & $224 \times 224$ \\
    \bottomrule
    \end{tabular}
}
\end{table}

\subsection{Fine-tuning}
\label{app:finetuning}

\begin{table}[H]
\centering
\caption{Hyperparameters for fine-tuning on ImageNet-1K.}
\label{tbl:ft:imagenet:hyperparams}
\small
\scalebox{0.95}{
    \begin{tabular}{l|c}
    \toprule
    \bf Hyperparameters & \bf ViT-B/16 \\
    \midrule
    Peak learning rate & 5e-4\\
    Epochs & 100\\
    Warmup epochs & 10\\
    Layer-wise learning rate decay & 0.65\\
    Batch size & 1024 \\
    Adam $\epsilon$ & 1e-8  \\
    Adam $\beta$ & (0.9, 0.999) \\
    Minimal learning rate & 4e-4 \\
    Learning rate schedule & Cosine \\
    \midrule
    Stoch. depth & 0.1 \\
    Repeated Aug & \xmark \\
    Weight decay & 0.05 \\
    Label smoothing $\varepsilon$ & 0.1     \\
    Dropout & \xmark \\
    Gradient clipping & \xmark \\
    \midrule
    Erasing prob.  & 0.25 \\
    Input resolution & $224 \times 224$ \\
    Rand Augment  & 9/0.5 \\
    Mixup prob.  & 0.6     \\
    Cutmix prob.  & 0.75    \\
    \bottomrule
    \end{tabular}
}
\end{table}

\begin{table}[H]
\centering
\caption{Hyperparameters for semantic segmentation on ADE20K (UperNet, 160K iterations).}
\label{tbl:ft:ade20k:hyperparams}
\small
\scalebox{0.95}{
    \begin{tabular}{l|c}
    \toprule
    \bf Hyperparameters & \bf ViT-B/16 \\
    \midrule
    Relative positional embeddings & \cmark \\
    Shared relative positional embeddings & \xmark \\
    \midrule
    Epochs & 50\\
    Peak learning rate & 0.5e-4 \\
    Fine-tuning steps & 160K \\
    Batch size & 16 \\
    Adam $\epsilon$ & 1e-8  \\
    Adam $\beta$ & (0.9, 0.999) \\
    Layer-wise learning rate decay & 0.75 \\
    Minimal learning rate & 0 \\
    Learning rate schedule & Linear \\
    Warmup steps & 1500 \\
    \midrule
    Dropout & \xmark \\
    Stoch. depth & 0.1 \\
    Weight decay & 0.05 \\
    \midrule
    Input resolution & $512 \times 512$ \\
    \bottomrule
    \end{tabular}
}
\end{table}

\section{Evolved Masking Hyperparameters}
\label{app:evolved_hyper}

\begin{figure*}[t!]
    \centering
    \includegraphics[width=\textwidth]{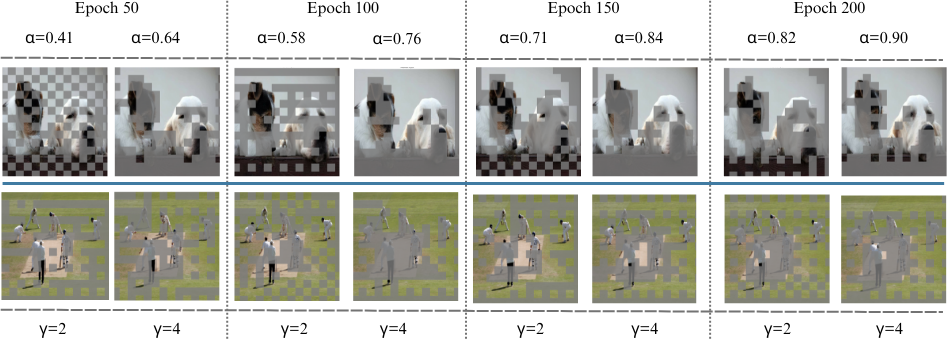}
    \caption{Comparison of HC evolved masking methods using $\gamma=2$ and $\gamma=4$ for scheduling the $\alpha$ values across different pre-training epochs. For each epoch, within each row, the first column represents the evolved mask generated using $\gamma=2$, and the second column shows the evolved mask using $\gamma=4$. Setting $\gamma=4$ slows down the transition, leading to more semantically rich masks later in training, thus making the task more challenging over time.}
    \label{fig:masks_gamma}
\end{figure*}

In our evolved masking strategy, we set specific hyperparameters to manage dynamic mask generation during pre-training. As shown in Table~\ref{tbl:mask:hyperparams}, we utilize an evolved mask type with a mask ratio of 0.75, meaning that 75\% of the image patches are masked in each training iteration. The transition from grid-based masking to attention-guided masking is governed by the gamma parameter ($\gamma$), set to 1.7. A higher $\gamma$ value slows down this transition, keeping the masking strategy closer to the initial grid-based masking for a longer period and transitioning to attention-guided masking later in the training. The position bias weight $\zeta$ is set to 0.9. For generating attention-guided masks, we employ Hierarchical Clustering (HC), with the number of clusters varying between 10 and 40 throughout training.

\begin{table}[H]
\centering
\caption{Evolved masking hyperparameters for pre-training on ImageNet-1K.}
\label{tbl:mask:hyperparams}
\small
\scalebox{0.98}{
    \begin{tabular}{l|c}
    \toprule
    \bf Hyperparameters & \bf ViT-B/16 \\
    \midrule
    Mask type & evolved \\
    Mask ratio & 0.75 \\
    Mask $\gamma$ & 1.7 \\
    Mask $\zeta$ & 0.9 \\
    Clustering Method & HC \\
    Min Clusters & 10 \\
    Max Clusters & 40 \\
    \bottomrule
    \end{tabular}
}
\end{table}

Figure~\ref{fig:masks_gamma} demonstrates the impact of different $\gamma$ values on the evolved masking process. By comparing $\gamma = 2$ and $\gamma = 4$, we observe that a higher $\gamma$ leads to the emergence of more semantically rich masks later in training. This delays the transition to attention-guided masking, making the reconstruction task gradually more challenging and fostering better feature learning over time.

\section{Loss Weight Sweep}
\label{app:loss_weight_sweep}

Table~\ref{tab:full_weight_sweep} provides the complete loss weight sweep data summarized in Figure~\ref{fig:weight_sweep} of the main paper. The decoder loss weight (DLW) controls the pixel reconstruction objective, while the CLS loss weight (CLW) controls the global alignment objective. The patch-level distillation weight is fixed at 1.0 throughout.

\begin{table}[H]
    \centering
    \caption{Full loss weight sweep. Top: pixel reconstruction weight (DLW) with no CLS loss. Middle: CLS alignment weight (CLW) with no pixel loss. Bottom: combined DLW and CLW.}
    \label{tab:full_weight_sweep}
    \resizebox{\linewidth}{!}{%
        \begin{tabular}{lcccccc}
            \toprule
            \textbf{Dec.\ Loss} & \textbf{DLW} & \textbf{CLW} & \textbf{kNN Top-1} & \textbf{kNN Top-5} & \textbf{Best Epoch} & \textbf{Fine-tune}\\
            \midrule
            \multicolumn{7}{c}{\textbf{Pixel Reconstruction Weight (DLW) Variations}} \\
            \midrule
            L2 & 0.50 & -- & 61.57 & 82.21 & 299 & 82.70 \\
            L2 & 0.40 & -- & 56.74 & 79.13 & 274 & 83.30 \\
            L2 & 0.30 & -- & 63.26 & 84.45 & 299 & 83.63 \\
            L2 & 0.20 & -- & 62.27 & 83.11 & 247 & 84.00 \\
            L2 & 0.10 & -- & 63.30 & 84.65 & 225 & 84.70 \\
            L2 & 0.05 & -- & 65.32 & 85.42 & 250 & 84.80 \\
            L2 & 0.01 & -- & 71.35 & 89.65 & 258 & 84.82 \\
            L2 & 0.005& -- & 54.19 & 77.27 & 258 & 84.64 \\
            \midrule
            \multicolumn{7}{c}{\textbf{CLS Alignment Weight (CLW) Variations}} \\
            \midrule
            -- & -- & 0.50 & 71.10 & 90.29 & 282 & 83.35 \\
            -- & -- & 0.30 & 72.33 & 91.11 & 288 & 83.78 \\
            -- & -- & 0.20 & 56.05 & 79.63 & 259 & 83.80 \\
            -- & -- & 0.10 & 57.93 & 80.57 & 275 & 84.86 \\
            \midrule
            \multicolumn{7}{c}{\textbf{Combined DLW + CLW}} \\
            \midrule
            L2 & 0.01 & 0.30 & \textbf{73.92} & \textbf{92.14} & 250 & \textbf{85.07} \\
            L2 & 0.10 & 0.10 & 73.88 & 92.20 & 299 & 84.17 \\
            \bottomrule
        \end{tabular}
    }
\end{table}

The DLW sweep reveals a sharp optimum at 0.01: increasing to 0.50 drops kNN by nearly 10 points, while decreasing to 0.005 causes a catastrophic drop to 54.19\%. The CLW sweep shows a similarly narrow effective range, with 0.30 achieving 72.33\% while 0.20 drops to 56.05\%. The combined optimum (DLW=0.01, CLW=0.30) achieves the best results across both kNN and fine-tuning metrics.

\section{Evolved Masking Sweep}
\label{app:masking_sweep}

Table~\ref{tab:full_masking_sweep} presents the complete set of evolved masking configurations evaluated, including variations in clustering method (EM vs.\ HC), transition rate ($\gamma$), position bias weight ($\zeta$), and mask ratio. Block masking serves as the baseline.

\begin{table}[H]
    \centering
    \caption{Full evolved masking sweep across 26 configurations. BM: block masking, EM: expectation-maximization clustering, HC: hierarchical clustering with relative position bias.}
    \label{tab:full_masking_sweep}
    \resizebox{\linewidth}{!}{%
        \begin{tabular}{lcccccc}
            \toprule
            \textbf{Method} & {$\boldsymbol\gamma$} & {$\boldsymbol\zeta$} & \textbf{Mask Ratio} & \textbf{kNN Top-1} & \textbf{kNN Top-5} & \textbf{Best Epoch} \\
            \midrule
            BM & -- & -- & 0.50 & \textbf{68.59} & \textbf{88.26} & 275 \\
            \midrule
            \multicolumn{7}{c}{\textbf{Mask Ratio = 0.50}} \\
            \midrule
            EM & 2.0 & 0.5 & 0.50 & 47.52 & 70.04 & 200 \\
            HC & 2.0 & 0.5 & 0.50 & 51.92 & 75.79 & 148 \\
            HC & 1.6 & 0.5 & 0.50 & 62.52 & 84.10 & 200 \\
            HC & 1.4 & 0.5 & 0.50 & 52.91 & 76.66 & 100 \\
            HC & 1.2 & 0.5 & 0.50 & 64.56 & 85.84 & 275 \\
            HC & 1.1 & 0.5 & 0.50 & 52.37 & 76.09 & 275 \\
            HC & 0.5 & 0.5 & 0.50 & 59.02 & 81.60 & 180 \\
            HC & 0.3 & 0.5 & 0.50 & 52.35 & 76.47 & 168 \\
            HC & 0.2 & 0.5 & 0.50 & 59.23 & 81.91 & 135 \\
            HC & 0.1 & 0.5 & 0.50 & 34.47 & 57.11 & 105 \\
            \midrule
            \multicolumn{7}{c}{\textbf{Mask Ratio = 0.75}} \\
            \midrule
            EM & 2.0 & 0.5 & 0.75 & 55.36 & 78.65 & 300 \\
            HC & 2.0 & 0.5 & 0.75 & 56.84 & 80.37 & 300 \\
            HC & 1.7 & 0.5 & 0.75 & 64.93 & 85.64 & 300 \\
            HC & 1.6 & 0.5 & 0.75 & 62.11 & 83.53 & 225 \\
            HC & 1.5 & 0.5 & 0.75 & 64.45 & 85.14 & 300 \\
            HC & 1.4 & 0.5 & 0.75 & 28.38 & 48.38 & 250 \\
            HC & 1.3 & 0.5 & 0.75 & 30.24 & 52.30 & 175 \\
            HC & 1.1 & 0.5 & 0.75 & 57.26 & 79.88 & 200 \\
            \midrule
            \multicolumn{7}{c}{\textbf{Position Bias Weight ($\zeta$) Variations at $\gamma$=1.5, Ratio=0.75}} \\
            \midrule
            HC & 1.5 & 0.3 & 0.75 & 21.10 & 39.90 & 175 \\
            HC & 1.5 & 0.4 & 0.75 & 38.46 & 62.60 & 200 \\
            HC & 1.5 & 0.5 & 0.75 & 64.45 & 85.14 & 300 \\
            HC & 1.5 & 0.6 & 0.75 & 56.13 & 79.69 & 200 \\
            HC & 1.5 & 0.7 & 0.75 & 63.02 & 83.92 & 285 \\
            HC & 1.5 & 0.8 & 0.75 & 63.03 & 84.34 & 280 \\
            HC & 1.5 & 0.9 & 0.75 & 64.16 & 85.12 & 250 \\
            \bottomrule
        \end{tabular}
    }
\end{table}

Across all 26 evolved masking configurations, no setting surpasses the block masking baseline of 68.59\%. The best HC result (64.93\% at $\gamma$=1.7, ratio=0.75) leaves a 3.7-point gap. HC consistently outperforms EM at matched settings (e.g., 51.92\% vs.\ 47.52\% at $\gamma$=2.0, ratio=0.50), confirming the value of relative position bias. The $\zeta$ sweep shows high sensitivity to the position bias weight, with values below 0.5 causing significant degradation.

\end{document}